\documentclass[10pt, a4paper]{article}
\usepackage{amssymb}
\usepackage{booktabs}
\usepackage{svg}
\usepackage{graphicx}
\usepackage{subcaption}
\usepackage{multirow}
\usepackage[final]{lrec2026}

\title{FalAR: A Large-scale Speaker-Annotated European Portuguese Speech Corpus of Parliamentary Sessions}

\name{\vspace{-0.5cm}\\
\textbf{Francisco Teixeira$^1$, Carlos Carvalho$^{1,2}$, Mariana Julião$^{1,2}$, Catarina Botelho$^1$, Rubén Solera-Ureña$^1$,}\\
\textbf{Sérgio Paulo$^1$, Thomas Rolland$^1$, Ben Peters$^1$, Isabel Trancoso$^{1,2}$, Alberto Abad$^{1,2}$}}

\address{$^1$INESC-ID, Lisbon, Portugal,\\ $^2$Instituto Superior T\'{e}cnico, Universidade de Lisboa, Portugal \\
         \{francisco.s.teixeira, carlos.carvalho\}@inesc-id.pt\\}

\abstract{
State-of-the-art performance for Automatic Speech Recognition (ASR) largely depends on the availability of large-scale labeled corpora. This creates a demand for increased data collection efforts, particularly for under-represented languages and dialectal varieties. Due to having considerably fewer speakers (around 11 million), European Portuguese (EP) is overshadowed by Brazilian Portuguese (BP) (around 200 million speakers) in currently available large-scale speech data resources, resulting in under-performing speech-based systems for EP users. To address this gap, and following similar data collection efforts for other languages, we present FalAR, a large-scale, speaker-annotated speech corpus of European Portuguese parliamentary sessions. Spanning approximately 20 years, FalAR comprises 5,800 hours of speech data. In addition, 4,850 hours have speaker identity annotations, for a total of 1,180 speakers with associated metadata including age, gender, political affiliation, and parliamentary role. The corpus was built using a state-of-the-art EP CAMÕES ASR model for transcription-reference alignment. In this paper, we describe the data collection process, together with the main characteristics of the FalAR corpus. Furthermore, we evaluate the trade-off between data quantity and alignment accuracy on ASR performance, with our experiments demonstrating that incorporating FalAR as pre-training data yields up to 14\% relative WER improvement over baseline models.
\\ \newline \Keywords{parliamentary speech corpus, automatic speech recognition, speaker recognition} }

\begin{document}

\maketitleabstract

\section{Introduction}

Recent advances in Automatic Speech Recognition (ASR) have been driven by a combination of architectural innovations \cite{first_speech_transformer, rnn_vs_transformers, conformer, e_branchformer, fast_conformer}, increased computational power, and the growing availability of large-scale labeled speech corpora \cite{chan2021speechstew, whisper, libriheavy, owsm-v4}. However, this trend inherently benefits majority languages or those with abundant online resources, while disadvantaging those with limited data (i.e., low-resource languages)~\cite{chen2024robustspeechrepresentationlearning}. Furthermore, even among seemingly high-resource languages, certain dialects or regional varieties may remain under-represented. Varieties can differ substantially in pronunciation and vocabulary, causing models trained on the dominant variety to underperform when applied to others.

An example of this asymmetry can be found between European Portuguese (EP) and Brazilian Portuguese (BP), which are rarely differentiated in currently available speech databases, and are often treated as a completely homogeneous language.
Therefore, speech recognition systems trained on large-scale web-scraped Portuguese data predominantly observe BP speech due to its substantially larger speaker base -- approximately 197 million out of a total of 240 million Portuguese speakers -- causing these models to perform suboptimally when applied to EP or other Portuguese varieties \cite{carvalho2025camoes}.
A plausible explanation for this suboptimal performance lies in the phonetic, prosodic, syntactic and phonological differences between the two varieties.
One of the most striking differences between BP and EP concerns vowel reduction, which is much more extreme in EP than in BP, but there are also notable differences at the syntactic and lexical levels \cite{mateus2000phonology, rouas2008language, abad2009porting}.

Even though EP has a much smaller speaker base than BP, it is nonetheless the native variety of around 11 million speakers. Ensuring fair and equitable access to speech technologies by EP speakers therefore demands the collection of large-scale resources for this variety. 
Historically, labeled resources for EP have been limited in scale, with the largest datasets falling short of 100 hours of speech data \cite{neto1997design, trancoso2003evaluation, hagen2003hmm}.
To date, one of the largest reported combinations of transcribed EP datasets corresponds to the training data of the CAMÕES models \cite{carvalho2025camoes}, amounting to only 425 hours of speech. 

In this work, we leverage publicly available, manually annotated recordings of parliamentary sessions of the Portuguese parliament\footnote{\url{https://www.parlamento.pt}} to address this shortage.
Specifically, we introduce FalAR,\footnote{The full corpus is available at \url{https://huggingface.co/datasets/inesc-id/FalAR}.}$^,$\footnote{In Portuguese, "falar" means "to speak"; AR is the acronym of "Assembleia da República", the official name of the Portuguese parliament.} a large-scale, speaker-annotated speech corpus containing 5,800 hours of transcription-reference aligned speech that spans approximately 20 years of parliamentary sessions of the Portuguese parliament. Alongside speech recordings and transcriptions, FalAR also includes detailed speaker identity and associated metadata, including age, gender, political affiliation and parliamentary role for 1,180 unique speakers, corresponding to more than 4,850 hours of speech.

We consider our work to be complimentary to the recently published EuroSpeech corpus \cite{pfisterer2025eurospeech}, which compiles speech data from 22 European parliaments, including the Portuguese parliament. The scale of EuroSpeech required its authors to apply a generic approach when collecting and annotating the data from each parliament, whereas our collection focused solely on the Portuguese parliament. 
This allowed us to annotate speaker identities and include speaker metadata, which EuroSpeech is lacking (resulting in speaker-dependent partitions), and to use the state-of-the-art EP ASR models of CAMÕES to generate pseudo-transcriptions for alignment, resulting in a larger and higher quality corpus.

In addition, we conduct a series of experiments that evaluate the performance of FalAR subsets with progressively higher alignment error rates, to analyze the trade-offs between data size and alignment accuracy, and their impact on training downstream ASR models for EP. When evaluated on the out-of-domain CAMÕES benchmark, performance of out-of-domain models trained with FalAR improves steadily as the amount of data -- and correspondingly, the alignment error rate -- increases, with results approaching the performance of baseline models trained with in-domain data. Moreover, employing FalAR as pre-training data prior to fine-tuning with in-domain speech allows for relative improvements of up to 14\% WER compared to models trained from scratch.

Overall, the main contributions of this work are the following:
\begin{itemize}
    \item We introduce FalAR, a large-scale, speaker-annotated speech corpus with 5,800 hours of aligned speech to address the shortage of speech resources for European Portuguese.
    \item We collect speaker identity and metadata, including age, gender, political affiliation and parliamentary role, for 4,850 hours of speech corresponding to 1,180 unique speakers, with longitudinal data spanning up to 20 years.
    \item We conduct several experiments to assess the trade-offs between data size and alignment accuracy, with our best models presenting a 14\% WER relative improvement over our domain-specific baseline.
\end{itemize}

The paper is organized as follows:
Section \ref{sec:relatedwork} presents the relevant related work; Section \ref{sec:falar} describes the data collection effort and the FalAR corpus; Sections \ref{sec:exp} and \ref{sec:results} detail this paper's experiments and results; finally, Section \ref{sec:conclusions} presents conclusions and plans for future work.

\section{Related Work}
\label{sec:relatedwork}

The development of speech resources for EP has always been closely linked to the development of EP ASR systems, as evidenced by early examples of data collection efforts for EP speech technologies. 
For instance, BD-PUBLICO \cite{neto1997design}, a corpus of 25 hours of read newspaper articles was collected to be used as the training data of an early large vocabulary hybrid Hidden Markov Model (HMM)/Deep Neural Network (DNN) system~\cite{neto98large}.
Similarly, the EP portion of SpeechDat~\cite{hagen2003hmm}, comprising 81 hours of narrow-band telephone speech, was collected as a part of an European-level collection of spoken language resources~\cite{hoge1997european} for the development of speech-based technologies.
Likewise, ALERT~\cite{trancoso2003evaluation}, a broadcast news corpus comprising 74 hours of speech, was collected for and used to train the hybrid HMM/DNN AUDIMUS system, developed to automatically transcribe broadcast news in EP \cite{AUDIMUS_BN,NetoASR2008}.

As ASR architectures became more data-demanding, research and development efforts in Portuguese progressively shifted towards BP, favoured by the availability of more extensive datasets resulting from its considerably larger speaker population~\cite{alencar2008lsf,candido2023coraa,limaintelligentsystems2025,leal2025mupe}.

Contrarily, efforts for EP ASR became increasingly reliant on combinations of multiple datasets and small manually identified EP subsets of larger Portuguese corpora.
For instance \citet{Carvalho2021mscthesis,carvalho21_iberspeech} combined data from BD-PÚBLICO, ALERT and SpeechDat, amounting to close to 150 hours to train the first end-to-end Connectionist-Temporal-Classification (CTC)-Attention ASR model for EP. 
\citet{Campinho2021mscthesis} curated a similar amount of EP data (150 hours), leveraging 2 hours of manually identified EP speech from Common Voice~\citep{commonvoice}, 70 hours from an in-house corpus sourced from the Portuguese RTP channel and 80 hours of Fala Bracarense~\cite{falabracarense}, a corpus of socio-linguistic interviews, built for the study of the regional accent of the Braga region in Portugal. 
\citet{MouraodeSa2021mscthesis} collected a total of 54 hours of EP speech, using around 33 hours from Europarl-ST~\cite{europarl_st} (a speech translation corpus of recordings from the European Parliament), and 21 hours from Multilingual-TEDx (a corpus of TEDx talks).

However, the models resulting from the aforementioned collection efforts were consistently outperformed by baseline hybrid HMM/DNN systems trained using the same data, evidencing the impact that the lack of resources has had in the development of end-to-end ASR systems for EP.

More recently, \citet{carvalho2025camoes} introduced CAMÕES,\footnote{\url{https://huggingface.co/datasets/inesc-id/camoes_asr}} an ASR framework for EP. It consists of an evaluation benchmark with 46 hours of EP data and a collection of state-of-the-art ASR models trained/fine-tuned with 425 hours of EP speech, compiled from a mix of proprietary corpora and publicly available sources, spanning multiple domains and demographic groups.
This work was the first successful attempt to achieve EP results on par with the state-of-the-art for BP with end-to-end systems. 

Portuguese speech subsets are also present in several well-known large-scale multilingual corpora. However, in most cases, the BP and EP are not differentiated. 
More importantly, BP is significantly more prevalent in these corpora. Corpora comprising EP speech include the above-mentioned CommonVoice~\cite{commonvoice}, with close to 2 hours of EP, Multilingual LibriSpeech (MLS) \cite{Pratap2020MLSAL}, with around 56h of EP speech, and MuAViC \cite{muavic}, reaching close to 20 hours of EP.\footnote{EP hours were determined by manual inspection and annotation.}
The YODAS dataset \citep{li2023yodas, owsm-v4} and the MOSEL collection \cite{gaido-etal-2024-mosel} include over 20,000h of Portuguese speech, sourced predominantly from YouTube, with a large proportion unlabeled or automatically labeled.
However, the actual proportion of EP speech is unknown.

An increasingly common approach for creating large-scale speech and text resources has been the collection of recordings of parliamentary debates, which are often publicly available and accompanied by high-quality -- although not always fully-verbatim -- transcripts.

Examples of text-only resources include ParlaMint~\citep{erjavec2023parlamint}, a corpus of transcripts of parliamentary proceedings of 26 national European parliaments, and Europarl \citep{koehn-2005-europarl}, a translation corpus that includes parallel text from the European Parliament in 11 languages. Textual corpora from the Portuguese parliament have already been released in PTPARL-D~\citep{almeida2021ptparl} and ParlaMint-PT~\citep{aires2024compiling}, with transcriptions of debates spanning 1976 to 2019 and 2005 to 2019, respectively. 

In addition, there is a growing number of speech corpora compiled from parliamentary data. 
These include large-scale corpora, such as Europarl-ASR \cite{garces2021europarl} for English as well as corpora for under-represented languages such as Danish~\cite{kirkedal20_interspeech},  Catalan~\cite{kulebi-etal-2022-parlamentparla}, Norwegian~\cite{solberg-ortiz-2022-norwegian}, Croatian, Polish, and Serbian~\cite{ljubevsic2024parlaspeech}. 

To date, the multilingual VoxPopuli corpus, which contains recordings from the European Parliament representing all languages in the European Union, is the largest available resource for EP, with 17.5k hours of unlabeled EP speech that can be used for self-supervised learning (SSL)-based pre-training~\cite{mohamed2022self, zhang2023googleusmscalingautomatic}. Nevertheless, achieving robust performance in downstream tasks still requires labeled data that span a wide range of speech domains, age groups, and other relevant speech factors.

More recently, the EuroSpeech corpus \cite{pfisterer2025eurospeech} has released over 78k hours of labeled speech from 22 European national parliaments, including around 5,100 hours from the Portuguese parliament. Our work complements this work by focusing specifically on the collection of European Portuguese parliamentary speech. 
The use of the CAMÕES state-of-the-art ASR model allowed us to generate pseudo-transcripts for the alignment of speech and the parliamentary proceedings transcripts, resulting in higher quality annotations. Furthermore, we include manually verified speaker identity annotations and speaker metadata, information that is lacking from the EuroSpeech, making its train/dev/test partitions speaker-dependent.

\section{FalAR} 
\label{sec:falar}

\begin{figure*}[ht]
    \centering
    \includegraphics[width=\textwidth]{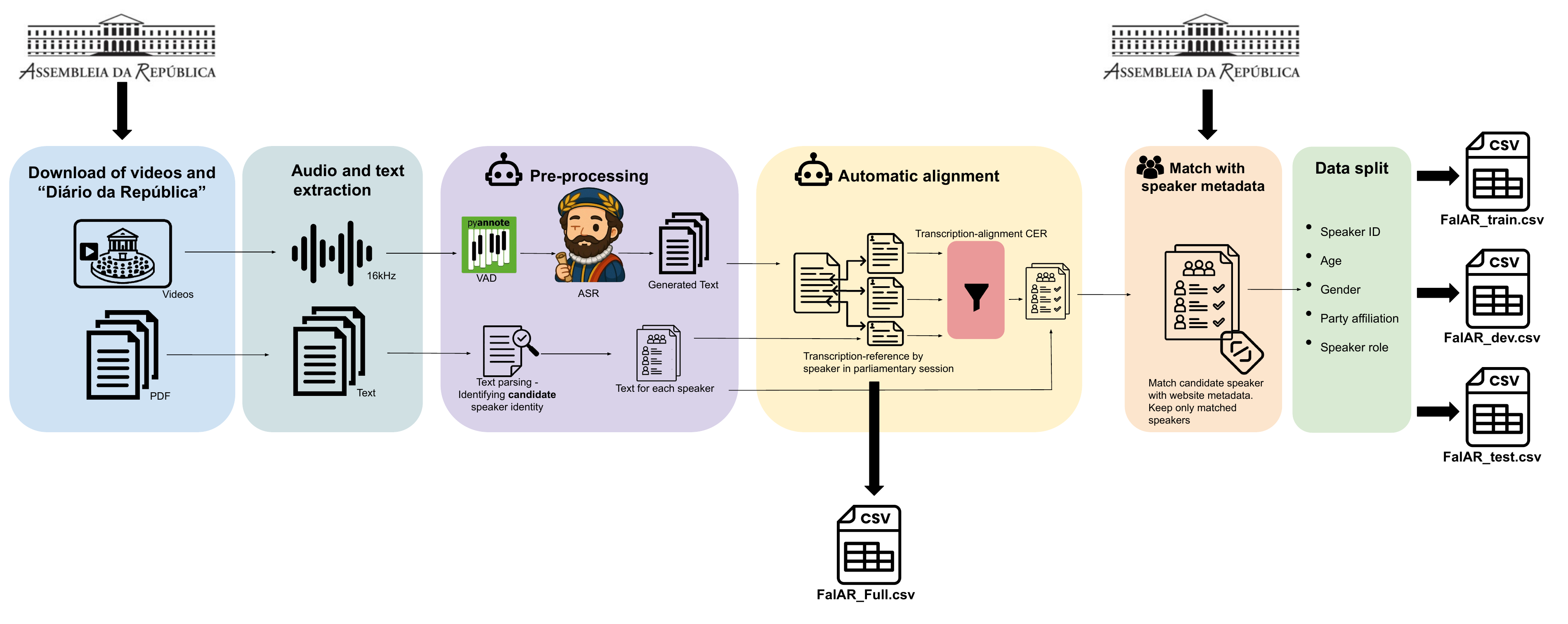}
    \caption{FalAR data collection and processing pipeline.
    }
    \label{fig:data_collection_pipeline}
\end{figure*}

The main objective of this work is to build a large-scale speech corpus for EP, leveraging the publicly available video recordings of Portuguese parliamentary meetings and corresponding manual transcriptions.

To achieve this, we collected the recordings, extracted and segmented the audio signals, and generated automatic transcriptions. These transcriptions were then aligned with the reference texts to determine the text segments corresponding to each utterance, which served as the ground-truth labels. For each utterance, we further identified a candidate speaker and matched them to the recording’s metadata to obtain reliable speaker labels. Finally, for every identified speaker, we manually annotated gender and date of birth (when this information was available online). The full data collection and processing pipeline is represented in Figure \ref{fig:data_collection_pipeline}.

The remainder of this section provides detailed description of the data collection, alignment, and annotation processes, as well as an overview of the resulting corpus. 

\subsection{Data collection}
\label{sec:falar-data-collection}
Since 2005, the Portuguese parliament has made its plenary sessions publicly accessible through its Audio-Visual Archive (AVA),\footnote{\url{https://av.parlamento.pt}} which contains complete session recordings as well as shorter clips of individual interventions. This archive is complemented by "Diário da Assembleia da República" (DAR)\footnote{\url{https://debates.parlamento.pt/}} -- the official gazette of the Portuguese parliament -- which provides detailed session summaries and full transcriptions (although not necessarily verbatim) of all interventions from each parliamentary meeting.

To construct our corpus, we first identified all available legislatures, legislative sessions, parliamentary meetings and individual interventions, and compiled a list of intervention video URLs. In parallel, we compiled a list of URLs for the DARs corresponding to each parliamentary meeting. This semi-automatic process resulted in the identification and successful download of 104,031 video interventions, along with their associated metadata (including speaker name, political affiliation, topic of intervention and parliamentary role) in HTML format, covering 2,041 parliamentary sessions over the last 20 years. For each of these sessions, we also retrieved the corresponding DARs, which include session summaries and full transcriptions of all interventions.

\subsection{Data pre-processing}

\subsubsection{Audio}
Audio was extracted from all downloaded videos, converted to 16-bit PCM format, and downsampled to 16 kHz. Each audio file was segmented into utterances of up to 30 seconds, using Pyannote's Voice Activity Detection (VAD) system \cite{Bredin23}. Subsequently, each segment was transcribed using CAMÕES's WhisperLv3-X\footnote{\url{https://huggingface.co/inesc-id/WhisperLv3-EP-X} a version of Whisper~\cite{whisper} fine-tuned on 425 hours of EP speech~\cite{carvalho2025camoes}.}.

\subsubsection{Text}
The downloaded DARs in HTML format were converted into plain text files, and each file underwent a semi-automatic filtering process to produce text files containing only the relevant content of speaker interventions (i.e., any spoken utterance by a participant during a parliamentary session that has been transcribed). 

Following this initial pre-processing stage, the next step involved identifying the beginning of each intervention, indicated by the pattern: "\{speaker name\}(\{affiliation\}): - ". 
Detecting this pattern enabled the extraction of the text between successive occurrences and its subsequent assignment to a candidate speaker, a necessary step to ensure correct speaker attribution.

However, manual inspection of initial groupings identified several cases in which speakers were designated by "O(A) Orador(a)" (i.e., \textit{the speaker}), a short-hand notation in DAR to indicate the main speaker of an intervention, when that speaker has been interrupted by another representative.
To correctly identify as many speakers as possible, we added a heuristic rule whereby, if the speaker is identified as "O(A) Orador(a)", the next-to-last speaker not labeled as "O(A) Orador(a)" is selected as the candidate speaker.

\subsection{Transcription-reference alignment}
After pre-processing the audio and text data, we aligned the automatic transcriptions with the reference DAR texts.
To accomplish this, we used the Smith-Waterman algorithm~\cite{smith_waterman}, which performs local sequence alignment by identifying the contiguous region within each reference text that most closely corresponds to a given transcription segment, accounting for partial matches and gaps.

To preserve speaker candidate information, each segment was aligned with portions of the reference text grouped by candidate speaker. For each segment, the aligned reference segment (and corresponding speaker) with the lowest Character Error Rate (CER) relative to the automatic segment transcription was selected as the gold-standard label.
This is done for two reasons: first, we consider that the segment from the reference text has a higher likelihood of being the true text; and second, selecting the reference text as the gold-standard label allows us to keep punctuation and capitalization, something that the automatically generated transcriptions do not contain.

\subsection{Speaker annotation}
The process described in the previous section resulted in the identification of 3,055 candidate speakers. 
Each speaker name was manually inspected to remove duplicate names due to typos, annotation errors and formatting issues. In addition, a number of speakers were identified only by their office (e.g., "minister", or "secretary of state"), in which case, they were annotated with the true speaker's name. This process resulted in 1,200 identified candidate speakers.

To minimize potential errors caused by alignment or formatting issues, we only considered as correct, candidate speakers whose names matched the speaker name contained in the videos' metadata.
It is important to note that this decision is particularly strict, since a large majority of videos contain interventions from more than one speaker, often including not only the speaker annotated in the video's metadata, but also a short intervention by the president of the parliament giving the floor to the speaker. Consequently, a large number of such interventions were left as non-verified speakers, i.e., stored without speaker annotation and left out of the speaker-independent partitions.

Following this stage, we proceeded to annotate each speaker with gender and date of birth, using online resources.
Out of the total 1,200 speakers, we were able to annotate the dates of birth (or, in a small number of cases, the year of birth) of 1,180 speakers. All identified speakers were annotated with gender information. 

To protect the privacy of the speakers present in this corpus to the best of our ability, we only provide numeric speaker identifiers, and omit the speaker's names. Moreover, we provide age annotations at the utterance level, instead of the speakers' dates of birth. 

\subsection{Corpus description}

\begin{table*}[t]
\setlength{\tabcolsep}{4pt}
\centering
\begin{tabular}{l ccccccccccc} 
\toprule
    & \multicolumn{2}{c}{\bf All data} & & \multicolumn{8}{c}{\bf Data with speaker age and gender information} \\
    \cmidrule{2-3}
    \cmidrule{5-12}
    & Duration & Word tokens & & \multicolumn{3}{c}{Duration (hours)} && \multicolumn{3}{c}{Speaker count} & Word Tokens \\
    \cmidrule{5-7} \cmidrule{9-11}
\bf CER & (hours) & (millions)  & & Total & M & F && Total & M & F & (millions) \\ 
\midrule
\bf <5\% & 1,503 & 13 && 1,366 & 923   & 443   && 1,159 & 772 & 387 & 12\\
\bf <10\% & 2,598& 23 && 2,370 & 1,621 & 749   && 1,173 & 783 & 390 & 21\\
\bf <15\% & 3,422& 31 && 3,122 & 2,153 & 969   && 1,177 & 787 & 390 & 28\\
\bf <20\% & 4,026& 36 && 3,664 & 2,539 & 1,125 && 1,178 & 788 & 390 & 33\\
\midrule
\bf Total&5,799&52&&4,852 & 3,399 & 1,453 && 1,180 & 790 & 390 & 44\\

\bottomrule
\end{tabular}
\caption{Dataset description, as a function of the CER threshold. \textit{M} and \textit{F} refer to male and female.
}
\label{tab:DatasetContentStats}
\end{table*}

The processes outlined in the previous sections yielded a corpus comprising 5,799 hours of transcribed speech data, of which 4,852 hours are annotated with speaker age and gender information, as summarized in Table~\ref{tab:DatasetContentStats}. 

For each audio utterance, we provide (1) the reference segment transcript obtained with the Smith-Waterman alignment algorithm, (2) the automatic transcription generated using CAMÕES’s WhisperLv3-X, (3) the CER between the two transcriptions, (4) the date of recording, (5) speaker id, (6) speaker gender, (7) speaker age, (8) speaker political affiliation, (9) speaker role -- e.g. \textit{Presidente da República} (President of the Republic), \textit{Deputado} (Member of parliament) --, and (10) intervention topic.

Table~\ref{tab:DatasetContentStats} also reports the size of the corpus across varying CER thresholds, obtained for the transcription-reference alignments. Lower CER thresholds indicate higher confidence in the accuracy of the aligned DAR transcription. 

\begin{figure*}[htbp]
    \centering
    \begin{subfigure}[b]{0.28\textwidth}
        \centering
        \includegraphics[width=\textwidth]{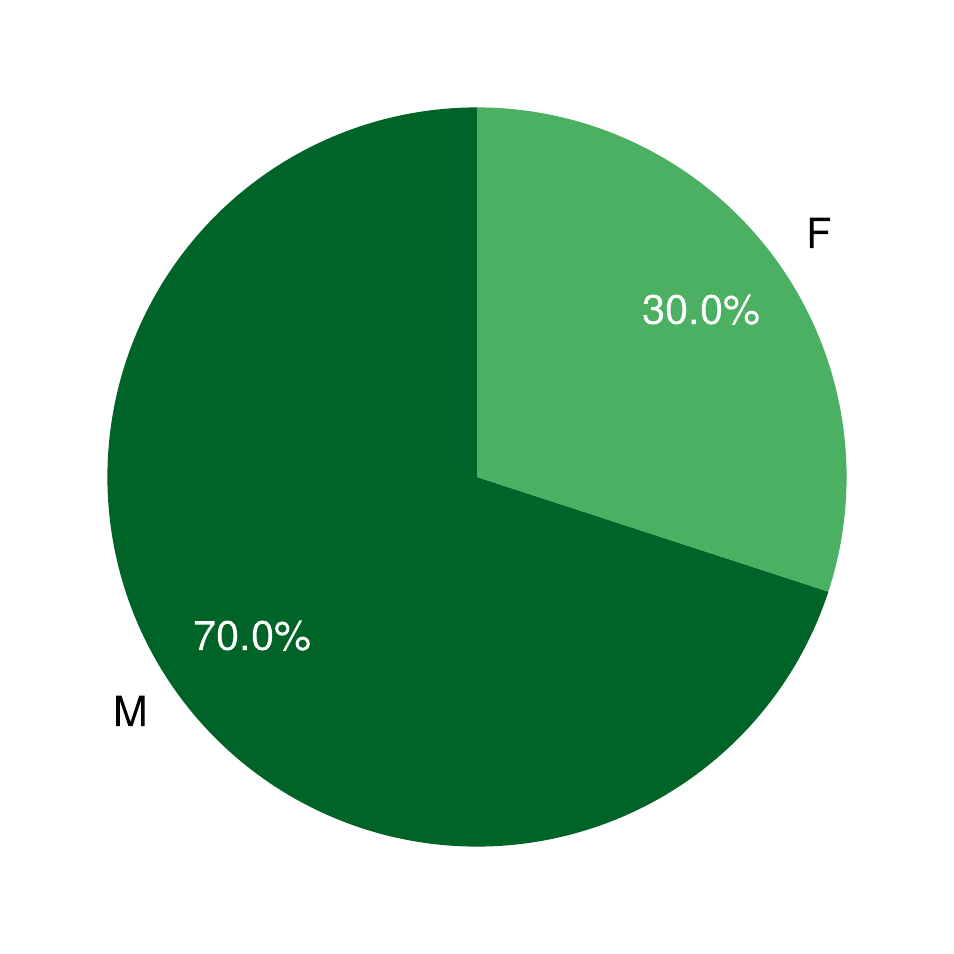}
        \caption{Gender}
        \label{fig:gender_distribution}
    \end{subfigure}
    \hfill
    \begin{subfigure}[b]{0.31\textwidth}
        \centering
        \includegraphics[width=\textwidth]{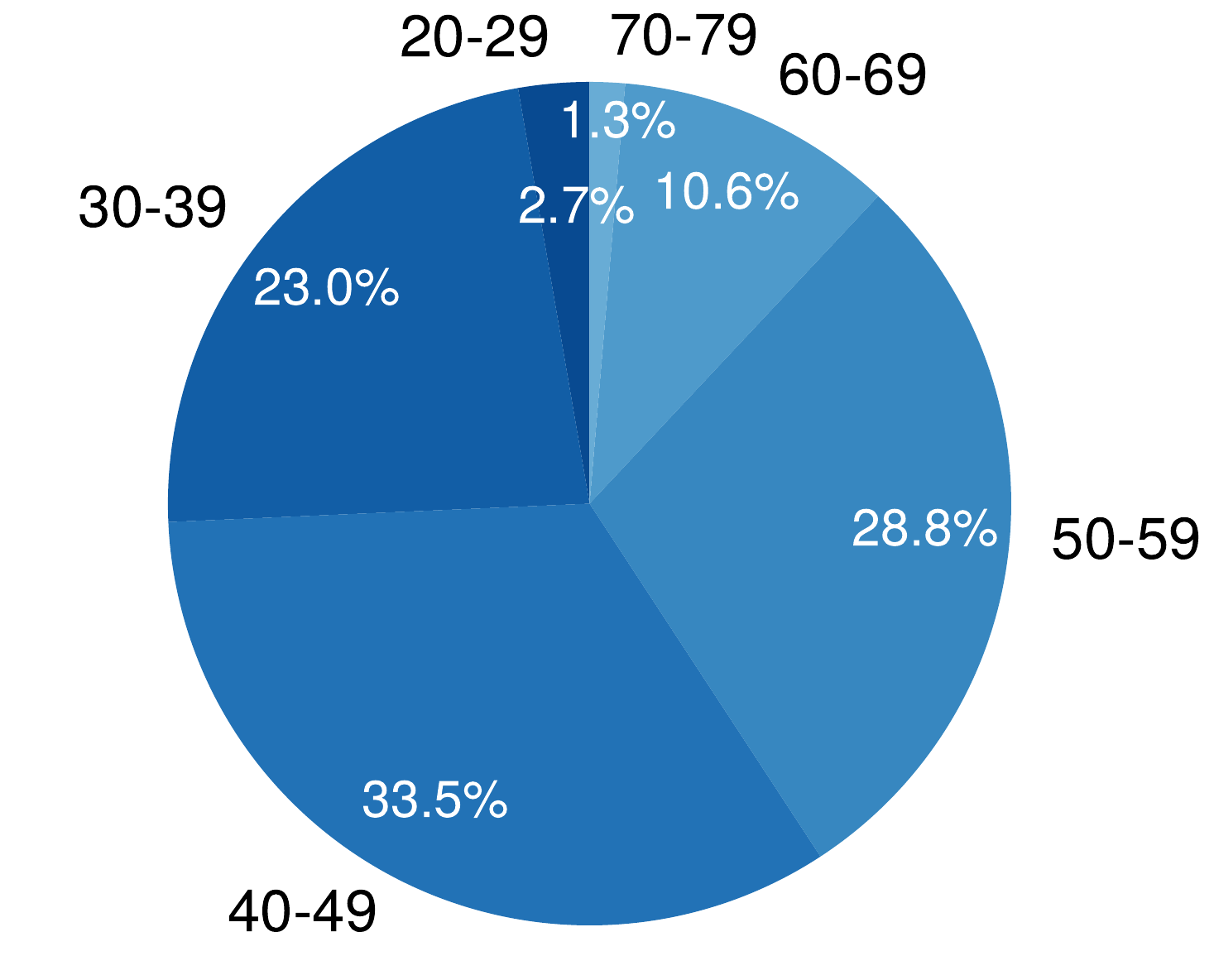}
        \caption{Age}
        \label{fig:age_distribution}
    \end{subfigure}
    \hfill
    \begin{subfigure}[b]{0.32\textwidth}
        \centering
        \includegraphics[width=\textwidth]{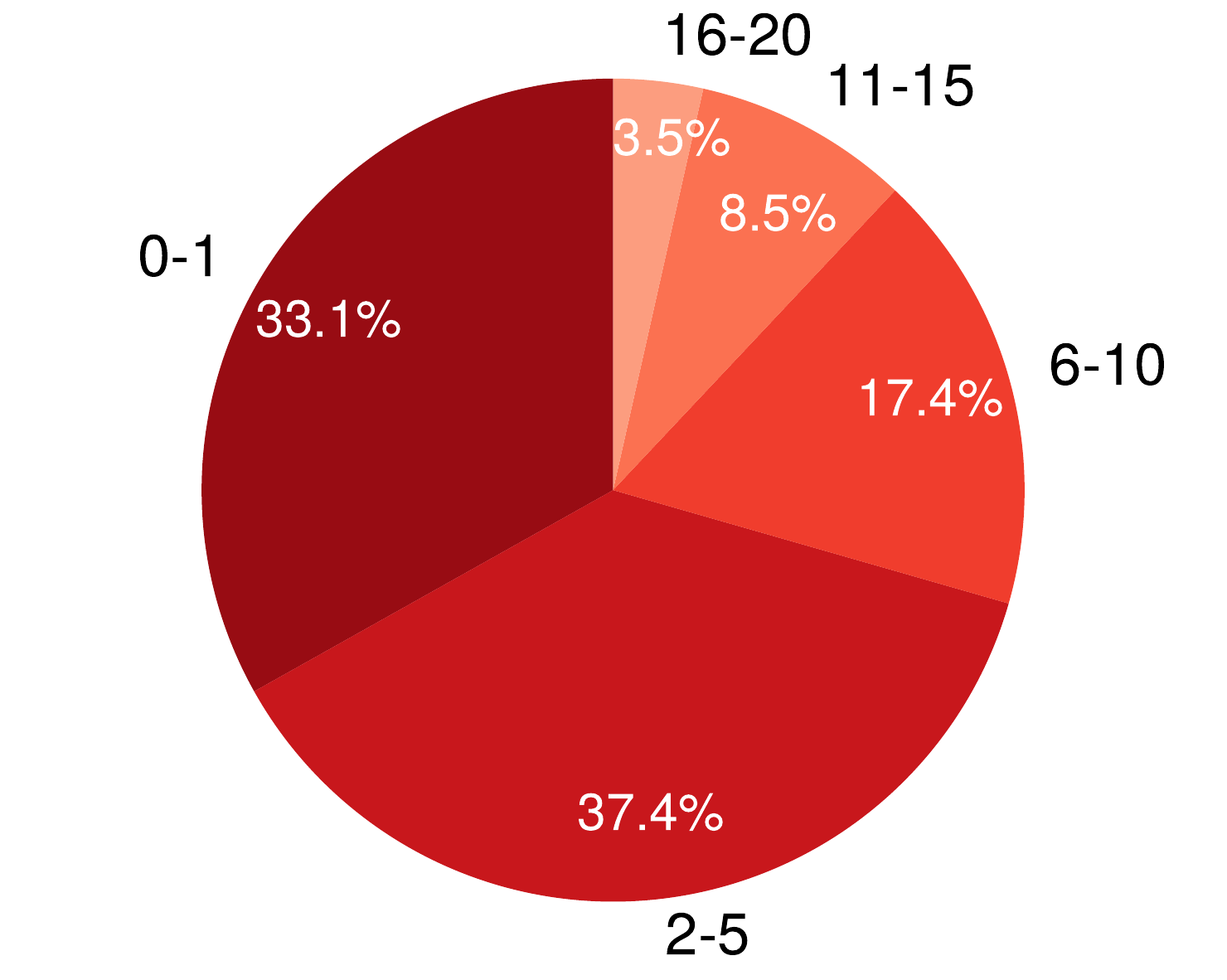}
        \caption{Longitudinal intervals}
        \label{fig:age_ranges}
    \end{subfigure}
    \caption{FalAR demographic distribution.}
    \label{fig:demographic_distribution}
\end{figure*}

The subset of the corpus containing speaker-level information includes 1,200 unique speakers, all of which have been annotated with gender information, while only 1,180 have been annotated with age information. Approximately 70\% of the speakers are male, while the remaining 30\% are female. The speakers' ages range from 20 to 79 years, with a predominantly middle-aged population. 
Each speaker contributes an average of four hours of speech. About 70\% of the speakers appear in recordings spanning up to five years, while approximately 3.5\% of the speakers have recordings that extend over a period of 16 to 20 years.
More detailed information regarding the demographic distribution of the corpus is presented in Figure \ref{fig:demographic_distribution}.

Besides making the entire dataset available (\textit{FalAR\_Full.csv} in Fig.~\ref{fig:data_collection_pipeline}), we also provide standardized speaker-independent train, development and test splits for the subset of the corpus containing speaker-level information, to promote reproducible research with this corpus. These partitions correspond to \textit{FalAR\_train.csv}, \textit{FalAR\_dev.csv}, and \textit{FalAR\_test.csv} in Figure ~\ref{fig:data_collection_pipeline} and comprise 4662/1171, 40/12, and 34/17 (hours/speakers), respectively.
The speakers in each partition are randomly and uniformly selected from the full set of annotated speakers.

Note that the test set is exclusively comprised of utterances with a maximum alignment CER of 5\%, to privilege high-quality transcriptions. This decision may make the test set less challenging, since utterances with lower alignment error rates may be inherently easier to transcribe, as they correspond to sentences where the model used to generate the transcriptions matched the reference text more closely. Nevertheless, we consider ensuring high-quality transcriptions in the test set to be more important than creating a more challenging but less reliable test set.

\section{Experimental Setup}
\label{sec:exp}

\subsection{Data}

To assess the impact of the proposed corpus on the performance of ASR models for European Portuguese, we conduct a series of experiments using different data configurations.

First, to determine the impact that different alignment error rates have in downstream ASR systems, we prepare five subsets of FalAR to train corresponding ASR models, as presented in Table \ref{tab:DatasetContentStats}, namely FalAR <\{5,10,15,20\}\% CER and FalAR <20\% CER + WL (all denoted by $FalAR_{<x\%}$ from hereon). 
The latter corresponds to the full corpus, where the labels for speech segments with CER $\geq$ 20\% correspond to the automatically generated transcripts (WL - weakly supervised), using the assumption that these are more reliable than the segments obtained by the alignment algorithm.
The data used in these five partitions is taken from \textit{All data} (cf. Table \ref{tab:DatasetContentStats}), for which speaker information may not necessarily be available, to maximize the quantity of data available.
The models trained with the aforementioned sets are all evaluated with in-domain data, using FalAR's test set (34.5h of speech with CER $<$ 5\%).

To explore the out-of-domain performance of the above-mentioned models, and to understand the impact of using FalAR as a pre-training corpus, followed by fine-tuning with data matching the evaluation domain, we used the corpus collected by the authors of CAMÕES~\cite{carvalho2025camoes}.
This dataset comprises over 470h of speech compiled from a set of 18 corpora from multiple domains -- 14 of these datasets are used for training and another intersecting set of 14 datasets is used in the test benchmark. 
The training set (denoted as EP-425) is 425 hours-long, whereas the test set is 46 hours-long, both comprising five domains, namely, read speech (RS), broadcast news (BN), talks and lectures (T/L), conversational speech (CS), and sociolinguistic interviews (SI).
An ASR model was additionally trained using solely CAMÕES, to provide a baseline with which to compare the FalAR-based models to.

It is important to note that, since we are using the \textit{All data} set for training, there may exist some speaker overlap with the test partition for non-speaker annotated samples. This is particularly true for $FalAR_{<20\%+WL}$, where the difference between the speaker annotated and non-speaker annotated sets is close to 1,000 hours of speech. 
This decision balanced the goal of understanding the impact of larger-scale data on out-of-domain performance against the computational cost of re-running the full set of experiments with our own speaker-independent partitions to better analyse the same effects on in-domain parliamentary data.

\begin{table*}[ht]
\centering
\resizebox{0.9\textwidth}{!}{
\begin{tabular}{l|l||r||r|r|r|r|r||r}
\toprule
\multicolumn{1}{c|}{\multirow{2}{*}{\textbf{Pre-training data}}} & \multirow{2}{*}{\textbf{Fine-tuning data}} & \multirow{2}{*}{\textbf{FalAR}} & \multicolumn{6}{c}{\textbf{CAMÕES}} \\\cmidrule{4-9}
& & & \multicolumn{1}{c|}{\textbf{RS}} & \multicolumn{1}{c|}{\textbf{BN}} & \multicolumn{1}{c|}{\textbf{T/L}} & \multicolumn{1}{c|}{\textbf{CS}} & \multicolumn{1}{c|}{\textbf{SI}} & \multicolumn{1}{c}{\textbf{Avg.}} \\
\midrule
$EP\mbox{-}425$ & -- & 15.5 &12.8 & 8.00& 21.2 & 22.1  & 39.1 & 20.6 \\
\midrule
$FalAR_{<5\%}$  & -- & 4.2 & 25.8 &13.6& 27.5 & 28.5 & 60.2 & 31.1 \\
$FalAR_{<10\%}$ & -- & 6.7 & 25.5 & 13.5 & 28.6 & 30.1 & 61.3 & 31.8 \\
$FalAR_{<15\%}$ & -- & 5.0 & 24.6 & 12.2 & 25.5 & 26.3 & 57.0 & 29.1 \\
$FalAR_{<20\%}$ & -- & 5.1 & 24.1 & 12.4 & 26.3 & 27.2 & 58.4 & 29.7  \\
$FalAR_{<20\%+WL}$ & -- & 3.1 & 19.0 & 9.9 & 20.9 & 22.2 & 50.0 & 24.4 \\
\midrule
$FalAR_{<5\%}$ & $EP\mbox{-}425$ & 13.9 & 11.1 & 7.8 & 20.5 & 20.0 & 37.3 & 19.3 \\
$FalAR_{<10\%}$ & $EP\mbox{-}425$ & 13.2 & 10.5 & 6.8 & 19.3 & 19.5 & 36.0 & 18.4 \\
$FalAR_{<15\%}$ & $EP\mbox{-}425$ & 12.8 & 9.4 & 6.6 & 18.2 & 18.8 & 35.7 & 17.7 \\
$FalAR_{<20\%}$ & $EP\mbox{-}425$ & 13.5 & 9.6 & 6.7 & 19.6 & 19.9 & 37.2 & 18.6 \\
$FalAR_{<20\%+WL}$ & $EP\mbox{-}425$ & 11.7  & 9.9 & 6.7 & 18.2 & 18.5 & 35.2 & 17.7 \\ 
\bottomrule
\end{tabular}
}
\caption{WER (\%) on FalAR test set and  CAMÕES benchmark. }
\label{tab:approaches}
\vspace{-0.2cm}
\end{table*}

\subsection{Implementation}
\label{sec:models}
We used the ESPnet toolkit \cite{watanabe2018espnet} for the core implementation and evaluation of our work. More specifically, we followed the LibriSpeech 960 \cite{7178964} recipe in ESPnet for training, decoding, and evaluation. 
All evaluated ASR models correspond to an E-Branchformer \cite{e_branchformer} with 144M trainable parameters, using 8x downsampling \citep{fast_conformer} and Flash Attention \cite{dao2022flashattention} to improve training and inference efficiency. The model's encoder comprises 17 layers, whereas the decoder is a 6-layer Transformer, both adapted from the original recipe. For the encoder module, we applied Rotary Positional Embeddings (RoPE) \cite{su2023roformerenhancedtransformerrotary}. We also adopted a piecewise-linear learning rate schedule \cite{owsm_ctc}, gradually increasing the learning rate from 2.0e-4 over the initial 15k steps to 2.0e-3 over the next 20k steps.

The training of the models with the different FalAR CER subsets in Table \ref{tab:DatasetContentStats} was carried out on a single NVIDIA H200 NVL GPU with 144GB of memory, using 120M batch bins and 15 epochs, except for $FalAR_{<20\%}$ and $FalAR_{<20\% + WL}$ models, which were trained for only 10 epochs. Training and fine-tuning of models with the CAMÕES corpus was performed on a single NVIDIA RTX 6000 Ada generation GPU with 48GB of memory, using 14M batch bins and 35 epochs. 

All models use the same text normalizer, based on the standard procedures used by Whisper, removing, among others, punctuation and capitalization, and expanding common acronyms. Performances are reported using WER (\%). Utterances shorter than 0.1 seconds or longer than 30 seconds were excluded during training and fine-tuning.

\section{Results}
\label{sec:results}

\subsection{In-domain performance}
Table \ref{tab:approaches} presents the in-domain results for the FalAR test set together with the out-of-domain performance on the CAMÕES benchmark, evaluated across its five domains. 

For the FalAR test set, we observe that performance generally improves as the training data size increases, with the largest gain and best overall performance of 3.1\% WER being caused by the addition of the WL data ($FalAR_{<20\%+WL}$ set). However, the second best result of 4.2\% WER comes from the $FalAR_{<5\%}$ set, worsening to 6.7\% when the $FalAR_{<10\%}$ set is used, and only then steadily dropping. These results for in-domain data suggest advantages of using either smaller training sets with high-quality alignments ($FalAR_{<5\%}$) or very large training sets, even with a lower quality ($FalAR_{<20\%+WL}$), compared to medium-sized training sets with intermediate transcription-reference alignment error levels. However, it is also possible that there exists a larger speaker overlap between the training and test data for the $FalAR_{<20\%+WL}$ subset that explains the strong performance of this model.

\subsection{Out-of-domain performance}

From the high-quality but smaller $FalAR_{<5\%}$ training set, to the larger $FalAR_{<20\%+WL}$ set, the average out-of-domain performance on CAMÕES improves from 31.1\% to 24.4\% WER. This proves that increasing the size of the FalAR training dataset allows the model to generalise better, even with a larger prevalence of transcription errors. 
Particularly, we observe that the addition of the WL data ($\sim$1800 additional hours) strongly contributes to the improvement of the out-of-domain performance of the model. When trained with this data, the model starts to become competitive with the baseline model trained with the CAMÕES EP-425 set, achieving comparable or even stronger results in the T/L and CS domains.
We hypothesize that sentences with higher alignment errors are harder for ASR models to transcribe than those with low alignment error rates due to worse and/or more varied acoustic conditions. Therefore, even though their transcriptions contain more errors, including these utterances in training helps the model generalize better to other domains.

\subsection{Fine-tuned performance}

The results obtained after fine-tuning the pre-trained models using 425 hours of speech data from the CAMÕES framework (EP-425) are presented in the lower half of Table~\ref{tab:approaches}. 

As anticipated, fine-tuning markedly improves performance on the CAMÕES benchmark relative to the out-of-domain condition.
Furthermore, all models incorporating pre-training show superior performance compared to the baseline model trained solely on EP-425, even though the pre-training data domain is highly specific (parliamentary speech) and differs substantially from those represented in the CAMÕES benchmark. 

Up to the 15\% CER threshold ($FalAR_{<15\%}$ set), larger amounts of pre-training data lead to better performance, with the best average result presenting a relative improvement of 14\% w.r.t. to the baseline model. However, beyond this point, model performance first degrades, and then stabilizes with the addition of the WL data. This is opposed to what was observed when evaluating the same models in the out-of-domain setting.
This may indicate that there is a maximum amount of pre-training data from the same domain from which models are able to benefit when fine-tuned for downstream domains. Alternately, as the model size remained fixed as we increased the amount of training data, the model pre-trained with $FalAR_{<20\%+WL}$ may simply not have the capacity to retain enough information about its additional pre-training data for it to still be beneficial after fine-tuning. The fact that the fine-tuned models' performance on FalAR's test set is markedly worse than the pre-trained models' performance on this data may provide evidence to this argument. Therefore, these results demand further experimentation to ascertain the exact causes of the reported performance.

\section{Conclusions}
\label{sec:conclusions}

This work introduces FalAR, to the best of our knowledge, the largest publicly available annotated European Portuguese speech corpus,  totalling 5,800 hours of parliamentary speech data. Our results show that using FalAR as pre-training data followed by in-domain fine-tuning improves ASR performance across all domains of the CAMÕES benchmark when compared to a strictly in-domain baseline model.

Although we provide standardized speaker-independent partitions together with this corpus, we do not provide results for the speaker-independent setup. We aim to address this in future work.

Furthermore, the proposed large-scale corpus offers two key advantages. First, it is expandable, as new recordings and corresponding transcripts from Portuguese parliamentary sessions are released continuously. Second, this corpus contains rich speaker metadata, including speaker identity, age, gender, and party affiliation, for 4,850 hours of speech from 1,180 speakers. These annotations do not only allow fair and controlled data splits for model development, but also longitudinal studies of speech and speaker traits spanning two decades. 
This dataset may thus serve as a valuable resource for research on ageing, charisma and other behavioural attributes, political science and possibly a wide range of unforeseen applications.

Future improvements to the FalAR corpus may also focus on providing baseline results for speaker recognition, punctuated and capitalized ASR models, improving high alignment CER transcriptions through weakly supervised methods, improving speaker labels through automatic speaker recognition, and releasing video data to support broader multimodal research. In a multimodal format, this resource could also potentially be turned into a searchable archive of parliamentary sessions~\cite{hansen2005speechfind}.

\section{Ethical considerations and limitations}

The source data that we curated and analysed to compile FalAR was obtained from publicly available open data resources (see Section~\ref{sec:falar-data-collection}).

In releasing the accompanying metadata, we deliberately omit personally identifiable information such as speaker names and dates of birth, and instead provide anonymised speaker identifiers and age information. Nevertheless, it must be acknowledged that, even if discarding biometric re-identification, complete anonymisation cannot be guaranteed, as re-identification is still possible by cross-referencing the provided metadata with publicly accessible information in the Portuguese Parliament website.
A further limitation of the data curation process rises from the extraction of speaker identifiers from PDF documents, which were frequently inconsistently formatted. These inconsistencies occasionally led to incomplete metadata entries. 

Finally, another limitation concerns the absence of punctuation marks in the automatically generated transcriptions, which may affect the quality of the alignment between the transcriptions and the reference texts. Nevertheless, the aligned reference transcriptions retain punctuation, offering an advantage for future research based on this corpus.

\section{Acknowledgements}
Work supported by Portuguese national funds through Fundação para a Ciência e a Tecnologia, I.P. (FCT) under projects UID/50021/2025 (DOI: \url{https://doi.org/10.54499/UID/50021/2025}) and UID/PRR/50021/2025 (DOI: \url{https://doi.org/10.54499/UID/PRR/50021/2025}) and by the Portuguese Recovery and Resilience Plan and NextGenerationEU European Union funds under project C644865762-00000008 (ACCELERAT.AI).

\section{Bibliographical References}\label{sec:reference}
\bibliographystyle{lrec2026-natbib}
\bibliography{lrec2026-example}

@inproceedings{aires2024compiling,
  title={Compiling and Exploring a {Portuguese} Parliamentary Corpus: {ParlaMint-PT}},
  author={Aires, Jos{\'e} and Cardoso, Aida and Pereira, Rui and Mendes, Am{\'a}lia},
  booktitle={Proc. of the IV Workshop on Creating, Analysing, and Increasing Accessibility of Parliamentary Corpora (ParlaCLARIN)@ LREC-COLING 2024},
  pages={12--20},
  year={2024}
}

@article{almeida2021ptparl,
  title={{PTPARL-D: an annotated corpus of forty-four years of Portuguese parliamentary debates}},
  author={Almeida, Paulo and Marques-Pita, Manuel and Gon{\c{c}}alves-S{\'a}, Joana},
  journal={Corpora},
  volume={16},
  number={3},
  pages={337--348},
  year={2021},
  publisher={Edinburgh University Press The Tun-Holyrood Road, 12 (2f) Jackson's Entry~…}
}

@INPROCEEDINGS{AUDIMUS_BN,
  author={Meinedo, Hugo and Souto, Nuno. and Neto, João P.},
  booktitle={Proc. ASRU}, 
  title={{Speech recognition of broadcast news for the European Portuguese language}}, 
  year={2001},
  volume={},
  number={},
  pages={319--322},
  doi={10.1109/ASRU.2001.1034651}
}

@inproceedings{carvalho21_iberspeech,
  title     = {{TRIBUS: An end-to-end automatic speech recognition system for European Portuguese}},
  author    = {Carlos Carvalho and Alberto Abad},
  year      = {2021},
  booktitle = {Proc. IberSPEECH},
  pages     = {185--189},
  doi       = {10.21437/IberSPEECH.2021-40},
}

@inproceedings{abad2009porting,
  title     = {{Porting an european portuguese broadcast news recognition system to brazilian portuguese}},
  author    = {Alberto Abad and Isabel Trancoso and Nelson Neto and M. Céu Viana},
  year      = {2009},
  booktitle = {{Interspeech 2009}},
  pages     = {92--95},
  doi       = {10.21437/Interspeech.2009-21},
  issn      = {2958-1796},
}

@inproceedings{carvalho2025camoes,
  title    = {{CAM{\~O}ES}: A Comprehensive Automatic Speech Recognition Benchmark for {European Portuguese}},
  author     = {Carvalho, Carlos and Teixeira, Francisco and Botelho, Catarina and Pompili, Anna and Solera-Ure{\~n}a, Rub{\'e}n and Paulo, S{\'e}rgio and Juli{\~a}o, Mariana and Rolland, Thomas and Mendon{\c{c}}a, John and Pereira, Diogo and Trancoso, Isabel and Abad, Alberto},
  booktitle = {Accepted to ASRU},
  year      = {2025}
}

@article{erjavec2023parlamint,
  title={{The ParlaMint corpora of parliamentary proceedings}},
  author={Erjavec, Toma{\v{z}} and Ogrodniczuk, Maciej and Osenova, Petya and Ljube{\v{s}}i{\'c}, Nikola and Simov, Kiril and Pan{\v{c}}ur, Andrej and Rudolf, Micha{\l} and Kopp, Maty{\'a}{\v{s}} and Barkarson, Starka{\dh}ur and Steingr{\'\i}msson, Stein{\th}{\'o}r and Çöltekin,  Çağrı and de Does, Jesse and Depuydt, Katrien and Agnoloni, Tommaso and Venturi, Giulia and Pérez, María Calzada and de Macedo, Luciana D. and Navarretta, Costanza and Luxardo, Giancarlo and Coole, Matthew and Rayson, Paul and Morkevičius, Vaidas and Krilavičius, Tomas and Darǵis, Roberts and Ring, Orsolya and van Heusden, Ruben and Marx,  Maarten and Fišer, Darja},
  journal={Language Resources and Evaluation},
  volume={57},
  number={1},
  pages={415--448},
  year={2023},
  publisher={Springer}
}

@article{garces2021europarl,
  title={{Europarl-ASR}: A large corpus of parliamentary debates for streaming {ASR} benchmarking and speech data filtering/verbatimization},
  author={Garc{\'e}s D{\'\i}az-Mun{\'\i}o, Gon{\c{c}}al and Silvestre Cerd{\`a}, Joan Albert and Jorge-Cano, Javier and Gim{\'e}nez Pastor, Adri{\'a}n and Iranzo-S{\'a}nchez, Javier and Baquero-Arnal, Pau and Rosell{\'o}, Nahuel and P{\'e}rez-Gonz{\'a}lez de Martos, Alejandro Manuel and Civera Saiz, Jorge and Sanchis Navarro, Jos{\'e} Alberto and Alfons, Juan},
  journal={Proc. Interspeech 2021},
  pages={3695--3699},
  year={2021},
  publisher={International Speech Communication Association (ISCA)}
}

@inproceedings{kirkedal20_interspeech,
  title     = { {FT Speech: Danish} Parliament Speech Corpus},
  author    = {Andreas Kirkedal and Marija Stepanović and Barbara Plank},
  year      = {2020},
  booktitle = {Proc. Interspeech},
  pages     = {442--446},
  doi       = {10.21437/Interspeech.2020-3164},
  issn      = {2958-1796},
}

@inproceedings{kulebi-etal-2022-parlamentparla,
    title = "{P}arlament{P}arla: A Speech Corpus of {C}atalan Parliamentary Sessions",
    author = "Kulebi, Baybars  and
      Armentano-Oller, Carme  and
      Rodriguez-Penagos, Carlos  and
      Villegas, Marta",
    editor = "Fi{\v{s}}er, Darja  and
      Eskevich, Maria  and
      Lenardi{\v{c}}, Jakob  and
      de Jong, Franciska",
    booktitle = "Proceedings of the Workshop ParlaCLARIN III within the 13th Language Resources and Evaluation Conference",
    month = jun,
    year = "2022",
    address = "Marseille, France",
    publisher = "European Language Resources Association",
    url = "https://aclanthology.org/2022.parlaclarin-1.18/",
    pages = "125--130"
}

@inproceedings{ljubevsic2024parlaspeech,
  title={The parlaspeech collection of automatically generated speech and text datasets from parliamentary proceedings},
  author={Ljube{\v{s}}i{\'c}, Nikola and Rupnik, Peter and Kor{\v{z}}inek, Danijel},
  booktitle={International Conference on Speech and Computer},
  pages={137--150},
  year={2024},
  organization={Springer}
}

@INPROCEEDINGS{NetoASR2008,
  author={Neto, João P. and Meinedo, Hugo and Viveiros, Márcio and Cassaca, Renato and Martins, Ciro and Caseiro, Diamantino},
  booktitle={Proc. ICASSP}, 
  title={{Broadcast news subtitling system in Portuguese}}, 
  year={2008},
  volume={},
  number={},
  pages={1561--1564},
  doi={10.1109/ICASSP.2008.4517921}
}

@inproceedings{solberg-ortiz-2022-norwegian,
    title = "The {N}orwegian Parliamentary Speech Corpus",
    author = "Solberg, Per Erik  and
      Ortiz, Pablo",
    editor = "Calzolari, Nicoletta  and
      B{\'e}chet, Fr{\'e}d{\'e}ric  and
      Blache, Philippe  and
      Choukri, Khalid  and
      Cieri, Christopher  and
      Declerck, Thierry  and
      Goggi, Sara  and
      Isahara, Hitoshi  and
      Maegaard, Bente  and
      Mariani, Joseph  and
      Mazo, H{\'e}l{\`e}ne  and
      Odijk, Jan  and
      Piperidis, Stelios",
    booktitle = "Proc. LREC",
    month = jun,
    year = "2022",
    address = "Marseille, France",
    publisher = "European Language Resources Association",
    url = "https://aclanthology.org/2022.lrec-1.106/",
    pages = "1003--1008"
}

@article{smith_waterman,
title = {Identification of common molecular subsequences},
journal = {Journal of Molecular Biology},
volume = {147},
number = {1},
pages = {195-197},
year = {1981},
issn = {0022-2836},
doi = {https://doi.org/10.1016/0022-2836(81)90087-5},
url = {https://www.sciencedirect.com/science/article/pii/0022283681900875},
author = {Temple F. Smith and Michael S. Waterman}
}

@article{rouas2008language,
title = {Language and variety verification on broadcast news for Portuguese},
journal = {Speech Communication},
volume = {50},
number = {11},
pages = {965-979},
year = {2008},
issn = {0167-6393},
doi = {https://doi.org/10.1016/j.specom.2008.05.006},
author = {Jean-Luc Rouas and Isabel Trancoso and Céu Viana and Mónica Abreu},
}

@inproceedings{watanabe2018espnet,
  title={Espnet: End-to-end speech processing toolkit},
  author={Watanabe, Shinji and Hori, Takaaki and Karita, Shigeki and Hayashi, Tomoki and Nishitoba, Jiro and Unno, Yuya and Soplin, Nelson Enrique Yalta and Heymann, Jahn and Wiesner, Matthew and Chen, Nanxin and Renduchintala, Adithya and Ochiai, Tsubasa},
  booktitle={Proc. Interspeech},
  year={2018},
}

@INPROCEEDINGS{7178964,
  author={Panayotov, Vassil and Chen, Guoguo and Povey, Daniel and Khudanpur, Sanjeev},
  booktitle={Proc. ICASSP}, 
  title={Librispeech: An {ASR} corpus based on public domain audio books}, 
  year={2015},
}

@inproceedings{dao2022flashattention,
  title={Flash{A}ttention: Fast and Memory-Efficient Exact Attention with {IO}-Awareness},
  author={Dao, Tri and Fu, Daniel Y. and Ermon, Stefano and Rudra, Atri and R{\'e}, Christopher},
  booktitle={Proc. NEURIPS},
  pages ={16344 - 16359},
  year={2022}
}

@INPROCEEDINGS{first_speech_transformer,
  author={Dong, Linhao and Xu, Shuang and Xu, Bo},
  booktitle={Proc. ICASSP}, 
  title={Speech-Transformer: A No-Recurrence Sequence-to-Sequence Model for Speech Recognition}, 
  year={2018},
  volume={},
  number={},
  pages={5884-5888},
  doi={10.1109/ICASSP.2018.8462506}}

@INPROCEEDINGS{rnn_vs_transformers,
  author={Karita, Shigeki and Chen, Nanxin and Hayashi, Tomoki and Hori, Takaaki and Inaguma, Hirofumi and Jiang, Ziyan and Someki, Masao and Soplin, Nelson Enrique Yalta and Yamamoto, Ryuichi and Wang, Xiaofei and Watanabe, Shinji and Yoshimura, Takenori and Zhang, Wangyou},
  booktitle={Proc. ASRU}, 
  title={A Comparative Study on Transformer vs {RNN} in Speech Applications}, 
  year={2019},
}

@inproceedings{conformer,
  author       = {Anmol Gulati and
                  James Qin and
                  Chung{-}Cheng Chiu and
                  Niki Parmar and
                  Yu Zhang and
                  Jiahui Yu and
                  Wei Han and
                  Shibo Wang and
                  Zhengdong Zhang and
                  Yonghui Wu and
                  Ruoming Pang},
  title        = {Conformer: Convolution-augmented Transformer for Speech Recognition},
  booktitle    = {Proc. Interspeech},
  year         = {2020},
}

@INPROCEEDINGS{e_branchformer,
  author={Kim, Kwangyoun and Wu, Felix and Peng, Yifan and Pan, Jing and Sridhar, Prashant and Han, Kyu J. and Watanabe, Shinji},
  booktitle={Proc. SLT}, 
  title={{E-Branchformer}: Branchformer with Enhanced Merging for Speech Recognition}, 
  year={2023},
}

@INPROCEEDINGS{fast_conformer,
  author={Rekesh, Dima and Koluguri, Nithin Rao and Kriman, Samuel and Majumdar, Somshubra and Noroozi, Vahid and Huang, He and Hrinchuk, Oleksii and Puvvada, Krishna and Kumar, Ankur and Balam, Jagadeesh and Ginsburg, Boris},
  booktitle={Proc. ASRU}, 
  title={Fast Conformer With Linearly Scalable Attention For Efficient Speech Recognition}, 
  year={2023},
  volume={},
  number={},
  pages={1-8},
  doi={10.1109/ASRU57964.2023.10389701}}

@article{chan2021speechstew,
  title={SpeechStew: Simply Mix All Available Speech Recognition Data to Train One Large Neural Network},
  author={Chan, William and Park, Daniel and Lee, Chris and Zhang, Yu and Le, Quoc and Norouzi, Mohammad},
  year={2021},
      eprint={2104.02133},
      archivePrefix={arXiv},
      primaryClass={cs.CL},
      journal={arXiv preprint, https://arxiv.org/abs/2104.02133}
}

@inproceedings{owsm-v4,
  title={{OWSM} v4: Improving Open Whisper-Style Speech Models via Data Scaling and Cleaning},
  author={Yifan Peng and Shakeel Muhammad and Yui Sudo and William Chen and Jinchuan Tian and Chyi-Jiunn Lin and Shinji Watanabe},
  booktitle={Proc Interspeech},
  year={2025},
}

@inproceedings{whisper,
author = {Radford, Alec and Kim, Jong Wook and Xu, Tao and Brockman, Greg and McLeavey, Christine and Sutskever, Ilya},
title = {Robust speech recognition via large-scale weak supervision},
year = {2023},
booktitle = {Proc. ICML},
articleno = {1182},
numpages = {27},
location = {Honolulu, Hawaii, USA},
}

@inproceedings{libriheavy,
      title={Libriheavy: a 50,000 hours ASR corpus with punctuation casing and context}, 
      author={Wei Kang and Xiaoyu Yang and Zengwei Yao and Fangjun Kuang and Yifan Yang and Liyong Guo and Long Lin and Daniel Povey},
      year={2024},
      booktitle={Proc. ICASSP}, 
}

@article{chen2024robustspeechrepresentationlearning,
      title={Towards Robust Speech Representation Learning for Thousands of Languages}, 
      author={William Chen and Wangyou Zhang and Yifan Peng and Xinjian Li and Jinchuan Tian and Jiatong Shi and Xuankai Chang and Soumi Maiti and Karen Livescu and Shinji Watanabe},
      year={2024},
      eprint={2407.00837},
      archivePrefix={},
      primaryClass={cs.CL},
      journal={arXiv pre-print, https://arxiv.org/abs/2407.00837}, 
}

@article{pfisterer2025eurospeech,
  title={EuroSpeech: A Multilingual Speech Corpus},
  author={Samuel Pfisterer and Florian Grötschla and Luca Lanzendörfer and Florian Yan and Roger Wattenhofer},
  journal={Proc. NeurIPS},
  year={2025}
}

@article{su2023roformerenhancedtransformerrotary,
title = {{RoFormer: Enhanced transformer with Rotary Position Embedding}},
journal = {Neurocomputing},
volume = {568},
pages = {127063},
year = {2024},
issn = {0925-2312},
doi = {https://doi.org/10.1016/j.neucom.2023.127063},
url = {https://www.sciencedirect.com/science/article/pii/S0925231223011864},
author = {Jianlin Su and Murtadha Ahmed and Yu Lu and Shengfeng Pan and Wen Bo and Yunfeng Liu},
}

@inproceedings{owsm_ctc,
    title = {{OWSM-CTC: An Open Encoder-Only Speech Foundation Model for Speech Recognition, Translation, and Language Identification}},
    author = "Peng, Yifan and Sudo, Yui and Shakeel, Muhammad and Watanabe, Shinji",
    booktitle = "Proc. ACL (Volume 1: Long Papers)",
    year = "2024",
    pages = "10192--10209"
}

@article{zhang2023googleusmscalingautomatic,
      title={{Google USM: Scaling Automatic Speech Recognition Beyond 100 Languages}}, 
      author={Yu Zhang and Wei Han and James Qin and Yongqiang Wang and Ankur Bapna and Zhehuai Chen and Nanxin Chen and Bo Li and Vera Axelrod and Gary Wang and Zhong Meng and Ke Hu and Andrew Rosenberg and Rohit Prabhavalkar and Daniel S. Park and Parisa Haghani and Jason Riesa and Ginger Perng and Hagen Soltau and Trevor Strohman and Bhuvana Ramabhadran and Tara Sainath and Pedro Moreno and Chung-Cheng Chiu and Johan Schalkwyk and Françoise Beaufays and Yonghui Wu},
      year={2023},
      eprint={2303.01037},
      archivePrefix={},
      primaryClass={cs.CL},
      journal={arXiv pre-print, https://arxiv.org/abs/2303.01037}, 
}

@article{mohamed2022self,
  title={{Self-supervised speech representation learning: A review}},
   author={Mohamed, Abdelrahman and Lee, Hung-yi and Borgholt, Lasse and Havtorn, Jakob D. and Edin, Joakim and Igel, Christian and Kirchhoff, Katrin and Li, Shang-Wen and Livescu, Karen and Maaløe, Lars and Sainath, Tara N. and Watanabe, Shinji},
  journal={IEEE Journal of Selected Topics in Signal Processing},
  volume={16},
  number={6},
  pages={1179--1210},
  year={2022},
  publisher={IEEE}
}

@book{mateus2000phonology,
    author = {Mateus, Maria Helena and d’Andrade, Ernesto},
    title = {The Phonology Of Portuguese},
    publisher = {Oxford University Press},
    year = {2000},
    isbn = {9780198235811},
    doi = {10.1093/oso/9780198235811.001.0001},
    url = {https://doi.org/10.1093/oso/9780198235811.001.0001},
}

@mastersthesis{Carvalho2021mscthesis,
  title        = {{TRIBUS: An end-to-end automatic speech recognition
 system for European Portuguese}},
  author       = {Carlos Carvalho},
  year         = 2021,
  month        = {January},
  address      = {Lisbon, Portugal},
  note         = {Available at \url{https://fenix.tecnico.ulisboa.pt/downloadFile/1126295043839127/81395-carlos-carvalho_dissertacao.pdf}},
  school       = {Instituto Superior Técnico, Universidade de Lisboa},
  type         = {Master's thesis}
}

@inproceedings{neto1997design,
  title={{The design of a large vocabulary speech corpus for Portuguese}},
  author={Neto, Jo{\~a}o P. and Martins, Ciro and Meinedo, Hugo and Almeida, Luis B},
  booktitle={Proc. Eurospeech},
  pages={1707--1710},
  year={1997},
}

@inproceedings{neto98large,
  title     = {A large vocabulary continuous speech recognition hybrid system for the portuguese language},
  author    = {Joao P. Neto and Ciro Martins and Luis B. Almeida},
  year      = {1998},
  booktitle = {5th International Conference on Spoken Language Processing (ICSLP 1998)},
  pages     = {paper 0562},
  doi       = {10.21437/ICSLP.1998-659},
  issn      = {2958-1796},
}

@inproceedings{hoge1997european,
  title={European speech databases for telephone applications},
  author={Hoge, Harald and Tropf, Herbert S and Winski, Richard and van den Heuvel, Henk and Haeb-Umbach, Reinhold and Choukri, Khalid},
  booktitle={Proc. ICASSP},
  volume={3},
  pages={1771--1774},
  year={1997},
  organization={IEEE}
}

@inproceedings{trancoso2003evaluation,
  title={{Evaluation of an alert system for selective dissemination of broadcast news}},
  author={Trancoso, Isabel and Neto, Joao P. and Meinedo, Hugo and Amaral, Rui},
  booktitle={Proc. Interspeech},
  pages={1257--1260},
  year={2003},
}

@mastersthesis{Campinho2021mscthesis,
  title        = {{Automatic speech recognition for European Portuguese}},
  author       = {Adriano Campinho},
  year         = 2021,
  month        = {July},
  address      = {Braga, Portugal},
  note         = {Available at \url{https://hdl.handle.net/1822/78249}},
  school       = {Escola de Engenharia, Universidade do Minho},
  type         = {Master's thesis}
}

@mastersthesis{MouraodeSa2021mscthesis,
  title        = {{Reconhecimento de fala em português de Portugal num contexto com poucos recursos}},
  author       = {João Manuel Alves {Mourão de Sá}},
  year         = 2021,
  month        = {November},
  address      = {Porto, Portugal},
  note         = {Available at \url{https://hdl.handle.net/10216/139258}},
  school       = {Faculdade de Ciências, Universidade do Porto},
  type         = {Master's thesis}
}

@inproceedings{commonvoice,
  title={{Common Voice: A Massively-Multilingual Speech Corpus}},
  author={Ardila, Rosana and Branson, Megan and Davis, Kelly and Kohler, Michael and Meyer, Josh and Henretty, Michael and Morais, Reuben and Saunders, Lindsay and Tyers, Francis and Weber, Gregor},
  booktitle={Proc. LREC},
  pages={4218--4222},
  year={2020}
}

@misc{falabracarense,
  title        = {{Perfil Sociolinguístico da Fala Bracarense}},
  author       = {{Centro de Estudos Humanísticos, Universidade do Minho}},
  howpublished = {\url{https://sites.google.com/site/projectofalabracarense/}},
  note         = {Accessed: 2025-10-24},
  year         = {2009}
}

@INPROCEEDINGS{europarl_st,
  author={Iranzo-Sánchez, Javier and Silvestre-Cerdà, Joan Albert and Jorge, Javier and Roselló, Nahuel and Giménez, Adrià and Sanchis, Albert and Civera, Jorge and Juan, Alfons},
  booktitle={Proc. ICASSP 2020}, 
  title={{Europarl-ST}: A Multilingual Corpus for Speech Translation of Parliamentary Debates}, 
  year={2020},
  volume={},
  number={},
  pages={8229-8233},
  doi={10.1109/ICASSP40776.2020.9054626}}

@inproceedings{Pratap2020MLSAL,
  title     = {MLS: A Large-Scale Multilingual Dataset for Speech Research},
  author    = {Vineel Pratap and Qiantong Xu and Anuroop Sriram and Gabriel Synnaeve and Ronan Collobert},
  year      = {2020},
  booktitle = {Proc. Interspeech},
  pages     = {2757--2761},
  doi       = {10.21437/Interspeech.2020-2826},
  issn      = {2958-1796},
}

@article{hansen2005speechfind,
  author={Hansen, J.H.L. and Rongqing Huang and Bowen Zhou and Seadle, M. and Deller, J.R. and Gurijala, A.R. and Kurimo, M. and Angkititrakul, P.},
  journal={IEEE Transactions on Speech and Audio Processing}, 
  title={SpeechFind: Advances in spoken document retrieval for a National Gallery of the Spoken Word}, 
  year={2005},
  volume={13},
  number={5},
  pages={712-730},
  doi={10.1109/TSA.2005.852088}}

@inproceedings{muavic,
  title     = {{MuAViC: A Multilingual Audio-Visual Corpus for Robust Speech Recognition and Robust Speech-to-Text Translation}},
 author    = {Mohamed Anwar and Bowen Shi and Vedanuj Goswami and Wei-Ning Hsu and Juan Pino and Changhan Wang},
  year      = {2023},
  booktitle = {Proc. Interspeech},
  pages     = {4064--4068},
  doi       = {10.21437/Interspeech.2023-2279},
  issn      = {2958-1796},
}

@inproceedings{koehn-2005-europarl,
    title = "{E}uroparl: A Parallel Corpus for Statistical Machine Translation",
    author = "Koehn, Philipp",
    booktitle = "Proceedings of Machine Translation Summit X: Papers",
    month = sep # " 13-15",
    year = "2005",
    address = "Phuket, Thailand",
    url = "https://aclanthology.org/2005.mtsummit-papers.11/",
    pages = "79--86",
}

@inproceedings{alencar2008lsf,
  title={{LSF and LPC-derived features for large vocabulary distributed continuous speech recognition in Brazilian Portuguese}},
  author={Alencar, Vladimir Fabregas Surigué de and Alcaim, Abraham},
  booktitle={Proc. 42nd Asilomar Conference on Signals, Systems and Computers},
  pages={1237--1241},
  year={2008},
}

@article{candido2023coraa,
  title={{CORAA ASR: a large corpus of spontaneous and prepared speech manually validated for speech recognition in Brazilian Portuguese}},
  author={Candido Junior, Arnaldo and Casanova, Edresson and Soares, Anderson and de Oliveira, Frederico Santos and Oliveira, Lucas and Junior, Ricardo Corso Fernandes and da Silva, Daniel Peixoto Pinto and Fayet, Fernando Gorgulho and Carlotto, Bruno Baldissera and Gris, Lucas Rafael Stefanel and Alu{\'i}sio, Sandra M.},
  journal={Language Resources and Evaluation},
  volume={57},
  pages={1139--1171},
  year={2023},
  publisher={Springer}
}

@InProceedings{limaintelligentsystems2025,
author="Lima, Rodrigo and Leal, Sidney E. and Junior, Arnaldo Candido and Alu{\'i}sio, Sandra M.",
title={A Large Dataset of Spontaneous Speech with the Accent Spoken in {S{\~a}o Paulo} for Automatic Speech Recognition Evaluation},
booktitle="Proc. Intelligent Systems: 34th Brazilian Conference (BRACIS)",
year="2025",
pages="33--47",
isbn="978-3-031-79029-4"
}

@inproceedings{leal2025mupe,
    title = "{M}u{P}e Life Stories Dataset: Spontaneous Speech in {B}razilian {P}ortuguese with a Case Study Evaluation on {ASR} Bias against Speakers Groups and Topic Modeling",
    author = "Evaldo Leal, Sidney  and
      Candido Junior, Arnaldo  and
      Marcacini, Ricardo  and
      Casanova, Edresson  and
      Gon{\c{c}}alves, Odilon  and
      Silva Soares, Anderson  and
      Freitas Lima, Rodrigo  and
      Stefanel Gris, Lucas Rafael  and
      Alu{\'i}sio, Sandra",
    editor = "Rambow, Owen  and
      Wanner, Leo  and
      Apidianaki, Marianna  and
      Al-Khalifa, Hend  and
      Eugenio, Barbara Di  and
      Schockaert, Steven",
    booktitle = "Proceedings of the 31st International Conference on Computational Linguistics",
    month = jan,
    year = "2025",
    address = "Abu Dhabi, UAE",
    publisher = "Association for Computational Linguistics",
    url = "https://aclanthology.org/2025.coling-main.407/",
    pages = "6076--6087"
}

@inproceedings{hagen2003hmm,
  title={{HMM/MLP hybrid speech recognizer for the Portuguese telephone SpeechDat corpus}},
  author={Hagen, Astrid and Neto, Joao P},
  booktitle={Proc. PROPOR},
  pages={126--134},
  year={2003},
}

@inproceedings{li2023yodas,
  title={YODAS: Youtube-Oriented Dataset for Audio and Speech},
  author={Li, Xinjian and Takamichi, Shinnosuke and Saeki, Takaaki and Chen, William and Shiota, Sayaka and Watanabe, Shinji},
  booktitle={Proc. ASRU},
  pages={1--8},
  year={2023},
}

@inproceedings{gaido-etal-2024-mosel,
    title = "{MOSEL}: 950,000 Hours of Speech Data for Open-Source Speech Foundation Model Training on {EU} Languages",
    author = "Gaido, Marco  and
      Papi, Sara  and
      Bentivogli, Luisa  and
      Brutti, Alessio  and
      Cettolo, Mauro  and
      Gretter, Roberto  and
      Matassoni, Marco  and
      Nabih, Mohamed  and
      Negri, Matteo",
    editor = "Al-Onaizan, Yaser  and
      Bansal, Mohit  and
      Chen, Yun-Nung",
    booktitle = "Proc. EMNLP",
    month = nov,
    year = "2024",
    address = "Miami, Florida, USA",
    publisher = "Association for Computational Linguistics",
    url = "https://aclanthology.org/2024.emnlp-main.771/",
    doi = "10.18653/v1/2024.emnlp-main.771",
    pages = "13934--13947",
}

@inproceedings{Bredin23,
  author={Hervé Bredin},
  title={{pyannote.audio 2.1 speaker diarization pipeline: principle, benchmark, and recipe}},
  year=2023,
  booktitle={Proc. Interspeech 2023},
}

\end{document}